\documentclass{article}
\usepackage{spconf,amsmath,graphicx}

\usepackage{cite}

\usepackage{enumitem}
\setlist{nosep, leftmargin=14pt}

\usepackage{mwe} % to get dummy images
\usepackage{booktabs}
\usepackage{multirow}

\title{NOA: a versatile, extensible tool for AI-based organoid analysis}
\name{Mikhail Konov$^{1,3,\star}$, Lion J. Gleiter$^{1,3,\star}$, Khoa Co$^{2}$, Monica Yabal$^{2}$, Tingying Peng$^{1,3}$}
\address{
$^{1}$ Helmholtz Munich, Munich, Germany \\
$^{2}$ Institute of Molecular Immunology,  School of Medicine and Health, TUM, Munich, Germany \\
$^{3}$ School of Computation, Information and Technology, TUM, Munich, Germany \\
$^{\star}$ \textit{equal contribution}
}
\begin{document}
%\ninept
%
\maketitle
\begin{abstract}

AI tools can greatly enhance the analysis of organoid microscopy images, from detection and segmentation to feature extraction and classification. However, their limited accessibility to biologists without programming experience remains a major barrier, resulting in labor-intensive and largely manual workflows. 
Although a few AI models for organoid analysis have been developed, most existing tools remain narrowly focused on specific tasks.

In this work, we introduce the Napari Organoid Analyzer (NOA), a general purpose graphical user interface to simplify AI-based organoid analysis. 
NOA integrates modules for detection, segmentation, tracking, feature extraction, custom feature annotation and ML-based feature prediction. 
It interfaces multiple state-of-the-art algorithms and is implemented as an open-source napari plugin for maximal flexibility and extensibility. 

We demonstrate the versatility of NOA through three case studies, involving the quantification of morphological changes during organoid differentiation, assessment of phototoxicity effects, and prediction of organoid viability and differentiation state. 
Together, these examples illustrate how NOA enables comprehensive, AI-driven organoid image analysis within an accessible and extensible framework.

\end{abstract}

\begin{figure*}[t]
\centering
\includegraphics[width=\textwidth]{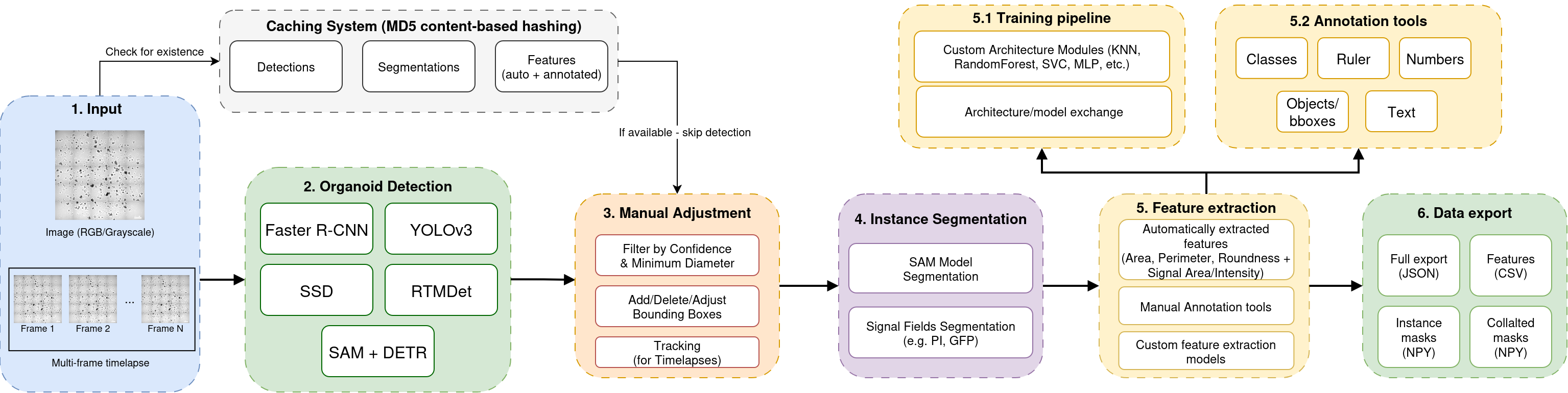}
\caption{NOA workflow: \textbf{1)} Loading of organoid images and timelapses. \textbf{2)} Detection using pre-trained models. Previous results are cached and can be restored. \textbf{3)} Filtering and manual editing of detections. For timelapses, tracking can align unique detection IDs across frames. \textbf{4)} Instance segmentation with SAM. Added signal images are also segmented. \textbf{5)} Computation of organoid features via \textbf{5.1)} ML models, trained with a built-in training pipeline or \textbf{5.2)} built-in annotation tools. \textbf{6)} Data export.}
\label{fig:workflow}
\end{figure*}

\section{Introduction}
\label{sec:intro}
Organoids are three-dimensional, stem cell–derived tissue models that recapitulate key structural and functional features of human organs. They have become indispensable tools in developmental biology, disease modeling, drug screening, and regenerative medicine, enabling experiments that bridge the gap between cell culture and in vivo systems~\cite{engineering_organoids}. 

When analyzing microscopy images of organoids, typical parameters of interest are their number and size \cite{organoID,orgaextractor,tellu}, their morphology \cite{organoID,tellu}, which may be categorized in multiple classes or quantified through a defined measurement such as roundness or wall thickness, fluorescent signals \cite{organoID,orgasegment}, and changes over time \cite{organoID,orgasegment,orgaextractor}.
AI has the potential to automate almost every step of the analysis pipeline, including detection, segmentation, feature extraction or other supervised prediction, enabling scalable and unbiased organoid analysis.

While the potential of AI tools to advance organoid image analysis has been demonstrated by multiple studies \cite{organoID,orgasegment,orgaquant,orgaextractor,tellu,bukas2024multiorg,cicceri2025}, the accessibility of the trained AI models to biologists remains limited. One major barrier is the lack of user-friendly implementations, as most algorithms do not provide a graphical user interface (GUI) and hence require substantial programming knowledge. 
Another limitation is the lacking standardization and interoperability between tools, making it difficult to integrate different analysis pipelines.

A further limitation lies in the specialization of existing tools, which are typically designed for specific analysis workflows or organoid types rather than general applicability. For example, Tellu detects and classifies mouse intestinal organoids into four pre-defined morphological classes \cite{tellu}, whereas OrgaSegment segments and tracks patient-derived organoids to measure organoid swelling \cite{orgasegment}. MultiOrg offers a napari-based GUI for organoid counting \cite{bukas2024multiorg}; however, it does not support segmentation, tracking, or feature extraction. 
To fully support the diversity of organoid research, a general-purpose tool should enable flexible workflows that are not restricted to a particular organoid model or task.

\textbf{Contribution.} In this work, we introduce the Napari Organoid Analyzer (NOA), a general-purpose graphical tool for organoid image analysis. NOA integrates all major analysis steps - detection, segmentation, tracking, feature extraction, annotation and ML-based prediction - within a single GUI, creating an easy-to-use but versatile tool. 

NOA automates analysis using state-of-the-art AI methods. It incorporates advanced algorithms for organoid detection \cite{bukas2024multiorg} and leverages the foundational Segment Anything Model (SAM) \cite{SAM} for segmentation. Users can annotate organoids with custom classes, length measurements, counts, or free-form text, and train ML models for feature prediction directly within the interface.

Designed for flexibility and open development, NOA is implemented in Python and built on the napari platform \cite{napari}, ensuring easy integration of new models in PyTorch or TensorFlow. The open-source design further allows to extend the GUI or adapt modules 
to specific experimental requirements. 

We demonstrate the versatility of NOA through three case studies covering morphological quantification, phototoxicity assessment, and phenotype prediction.

\section{Methods}
\label{sec:methods}
Figure~\ref{fig:workflow} demonstrates NOA's comprehensive workflow that consists of detection, manual adjustment \& filtering, segmentation, feature extraction,annotation, and prediction, and data export. The tool is capable of processing RGB, RGBA and grayscale images 
as well as timelapses.

\subsection{Detection}

NOA provides four pre-trained detection models from a recent study on AI-based organoid detection \cite{bukas2024multiorg}, based on the Faster R-CNN, YOLOv3, SSD, and RTMDet implementation in \texttt{mmdetection} \cite{mmdetection}. 
Additionally, it provides a custom detection model trained on a combination of publicly available organoid and microscopy datasets (data published in \cite{neurips_cellseg,bukas2024multiorg,orgasegment,orgaquant,tellu}). The custom model utilizes SAM \cite{SAM} features with a DETR-based \cite{detr} detection head. The list of built-in detection/segmentation methods can be extended with new models.

Detection is performed using sliding windows with 50\% overlap, supporting multi-scale inference with configurable window sizes and downsampling rates. Detected bounding boxes undergo non-maximum suppression ($IoU > 0.5$). Detection can be restricted to user-drawn regions of interest (ROIs) for targeted analysis.

\textbf{For timelapses:} Detection is performed independently on each frame, creating separate sets of bounding boxes with unique IDs. If desired, tracking can be applied to align detection IDs across frames using TrackPy \cite{trackpy} with configurable search radius and memory parameters. Missing detections could be optionally filled by copying bounding boxes from previous frames.

\subsection{Manual adjustment \& filtering}

Since no detection model will be 100 \% correct, NOA provides users the option to manually correct detections by deleting, modifying or adding bounding boxes. Further, users can filter detections via confidence threshold and minimum diameter sliders. 
Manual adjustment is supported through napari's shape editing tools. All bounding boxes (including manually added ones) are assigned a unique detection ID and stored in the internal storage. 

\subsection{Segmentation}

To obtain instance segmentation masks, NOA promts SAM \cite{SAM} with the detected bounding boxes. Resulting instance segmentation masks are stored internally as polygon contours for memory efficiency and displayed in the napari viewer.
Additionally, one or multiple signal images (e.g., propidium iodide (PI), green fluorescent protein (GFP)) can be segmented alongside the original image using the same bounding box set as a prompt for SAM.

\subsection{Feature Extraction \& Annotation}

\textbf{Geometric features} (area, perimeter, roundness) are automatically calculated from segmentation masks during the segmentation stage. Additionally, \textbf{texture and intensity-based features} (mean/total signal intensity within instance masks and \texttt{skimage.measure.regionprops} features) are computed per organoid for each attached signal image, including the original image. 

For \textbf{manual annotation}, NOA provides five annotation tools: text, number, classes (multi-label), object (bounding boxes), and ruler (draws and measures multiple multi-segment lines). Manual annotations can be performed on all or a selected set of organoids. The annotation process can be stopped at any point and resumed later, with all ongoing annotation states being saved in the tool's cache.

For \textbf{automatic prediction} of custom features, NOA implements common ML methods for training classification and regression models. They are implemented in a modular way and define configuration, training, and inference procedures through a standardized interface. Built-in architectures include KNN, Random Forest, AdaBoost, MLP, Gaussian Process, and SVC. Additional model architectures can be imported on demand as long as they implement the same configuration, training and inference interface. 

Feature extraction models can be trained either on data which is currently opened in NOA or on datasets in CSV files, for example exported data from previously annotated images. The trained model can then be used to annotate new features automatically. Trained models with their corresponding architecture module are saved in the tool's cache and can be exported/imported for cross-session use.

\subsection{Storage and export/import}

All organoid detections, features, and segmentation masks are stored together in the tool's detection ID-indexed internal storage. The tool further supports caching based on an image hash to restore the internal storage and trained feature extraction models from previous analysis sessions.

Users can export all data from the internal storage for specific organoids into a JSON file, which can be later imported back into the tool to restore the previous state. Additionally, all features including bounding boxes can be exported into CSV files. Segmentation masks are exported into NPY files.

\section{Results}
\label{sec:results}

\subsection{User Interface}

\begin{figure}[htb]
\centering
\includegraphics[width=\columnwidth]{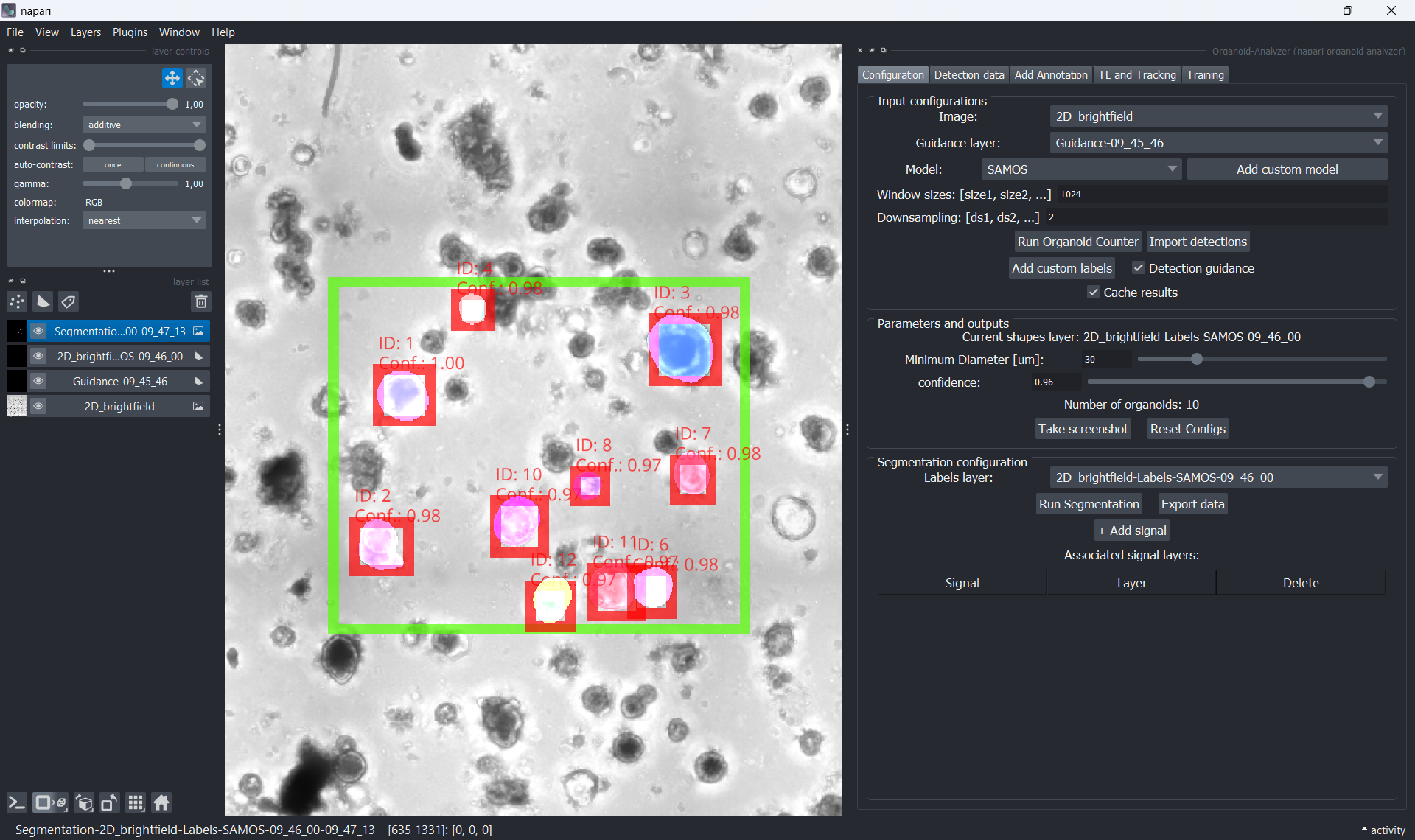}
\caption{Example of the NOA interface: napari viewer controls (left panel), interactive image view displaying detections/segmentation (center), and NOA's organoid analysis instruments (right panel). The image demonstrates detection and segmentation results (red boxes and multi-colored masks) within the user-defined ROI (green box).
}
\label{fig:NOA_UI}
\end{figure}

NOA provides a GUI (Figure~\ref{fig:NOA_UI}) that enables AI-based organoid analysis without programming knowledge. It interfaces state-of-the-art algorithms for organoid detection and segmentation. Additionally it enables tracking of individual organoids in timelapses and supports measuring and annotating diverse morphological features per organoid. Finally it implements a modular interface for training ML models to automate feature annotation.

\begin{figure}%[htb]
\centering
\includegraphics[width=\columnwidth]{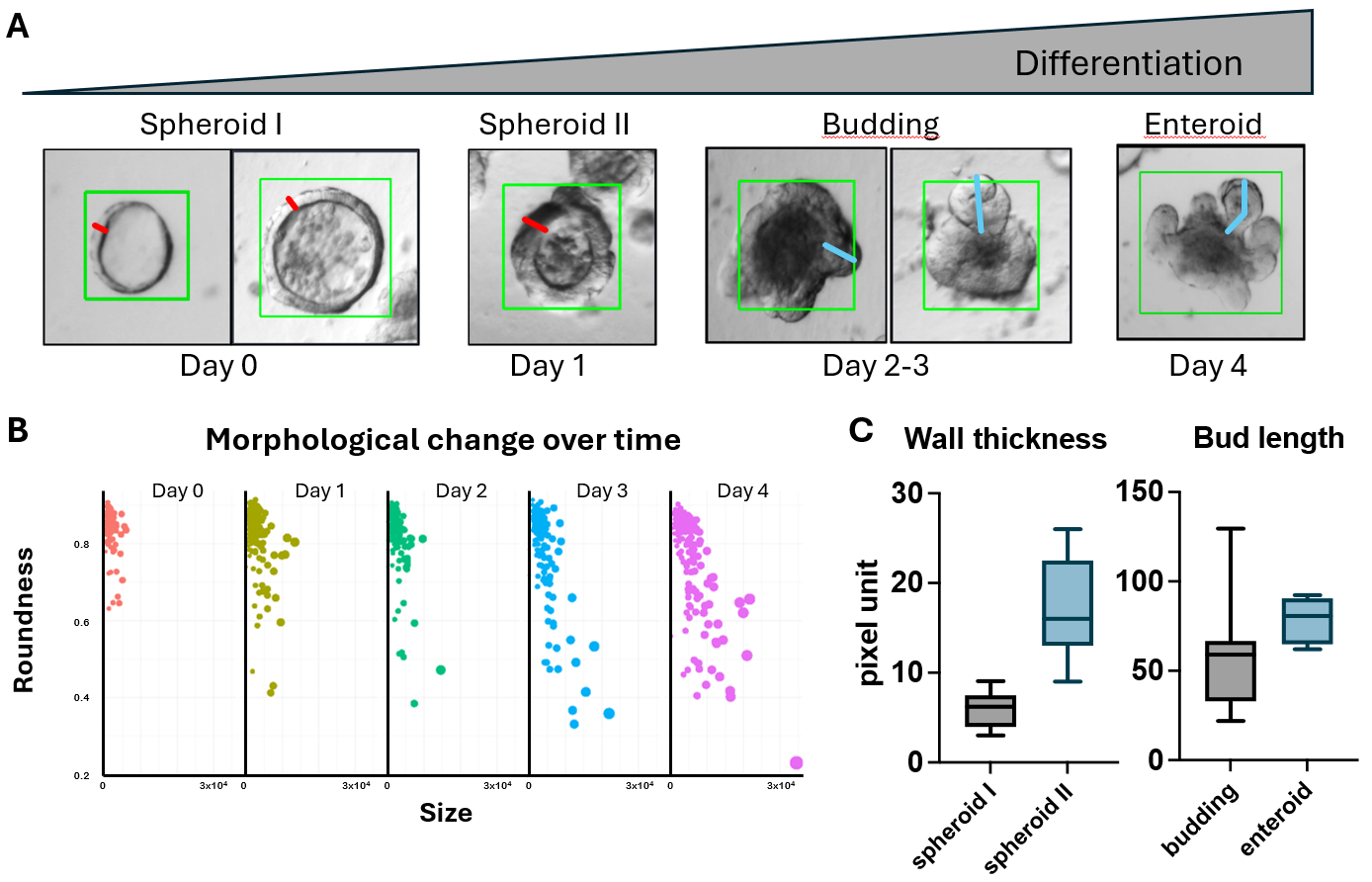}
\caption{Morphological changes of organoids due to differentiation. A) Examples of the annotated differentiation stages and features (red: wall thickness, blue: bud length). B) Geometric features (size and roundness) over time. C) Manually annotated features distinguish differentiation stages.}
\label{fig:experiments:morphology}
\end{figure}

\subsection{Biological Experiments}

We demonstrate the versatile functionality of NOA in three real-world experiments with mouse intestinal organoids, described in the following subsections.

Mouse small intestinal organoids (mSIOs) were maintained in 50\% Matrigel domes overlaid with IntestiCult™ Organoid Growth Medium (STEMCELL Technologies) and passaged every 4–5 days. For experimental assays and imaging, organoids were cultured in a low-viscosity system by resuspending them in medium containing 10\% Matrigel and plating them in 96-well plates. To synchronize differentiation, organoids were treated with CHIR99021 and Nicotinamide. Cell viability were monitored by detecting cleaved Caspase-3 activity using NucView® (Biotium). All images were acquired using a Leica Thunder microscope.

\subsubsection{Profiling organoid morphology}
\label{sec:results:bio:morphology}

Differentiating organoids are categorized as one of four differentiation stages: \textit{spheroid I}, \textit{spheroid II}, \textit{budding}, and \textit{enteroid}. 
We used NOA to compute geometric features such as size and roundness. Additionally we used NOA's ruler annotation function to measure wall thickness for spheroid I and spheroid II organoids and bud length for budding and enteroid organoids (Figure \ref{fig:experiments:morphology} A).

We observe a general trend of increasing size and decreasing roundness with progressing differentiation over time (Figure \ref{fig:experiments:morphology} B). 
Further, the manual annotation proves useful to characterize the morphological profile of organoids at different stages:
Spheroid I and spheroid II organoids show a clear difference in wall thickness, and budding and enteroid organoids show a clear difference in bud length (Figure \ref{fig:experiments:morphology} C).

\subsubsection{Measuring phototoxic effects}
\label{sec:results:bio:phototoxicity}

\begin{figure}[htb]
\centering
\includegraphics[width=\columnwidth]{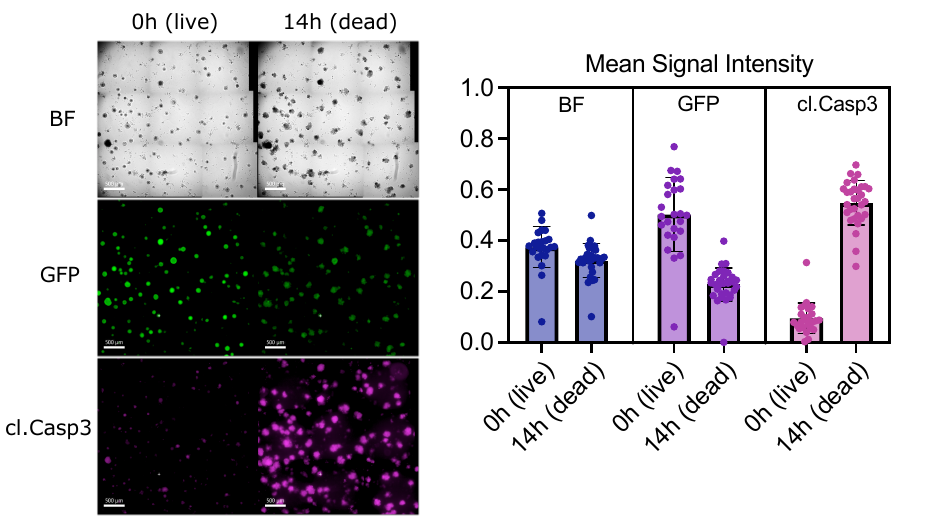}
\caption{Phototoxicity effect measured through mean signal intensity from brightfield, GFP, and cl.Casp3-stained images. Scalebars are 500 µm.}
\label{fig:experiments:phototoxicity}
\end{figure}

We demonstrate the utility of NOA for multi-channel microscopy images by comparing phototoxicity measurements from brightfield, GFP, and cleaved Caspase-3 (cl.Casp3)-stained images. We compare two time points: after 0 hours, where organoids are mostly alive, and after 14 hours, where organoids have mostly died from phototoxicity. While the brightfield image intensity changes little over time, mean GFP intensity (expressed in living cells) per organoid decreases and mean cl.Casp3-dye intensity (marks dead cells) increases substantially, reflecting the progressive loss of viability (Figure \ref{fig:experiments:phototoxicity}).

\subsubsection{Predicting organoid differentiation and viability}
\label{sec:results:bio:prediction}

\begin{table}%[htb]
\centering
\resizebox{\columnwidth}{!}{%
\begin{tabular}{lcc}
\toprule
Model\textbackslash Dataset & Differentiation & Viability \\
\midrule
MLP & \textbf{68.56 ± 10.22\%} & 71.21 ± 7.07\%
\\
SVC & 51.53  ± 13.40\% & 50.30 ± 13.15\% \\
Random Forest & 62.46 ± 8.39\% & \textbf{72.12 ± 6.33}\% \\
KNN & 55.65 ± 10.04\% & 70.30 ± 10.12\% \\
AdaBoost & 58.65 ± 10.23\% & 68.79 ± 5.08\% \\
Gaussian Process & 45.93 ± 7.53\% & 70.91 ± 6.24\% \\
\bottomrule
\end{tabular}%
}
\caption{Classification accuracy of ML models trained with NOA to predict differentiation stage and organoid viability. Mean and sd. using 10-fold cross-validation.}
\label{tab:ml_results}
\end{table}

We show how NOA can be used for training ML-models to predict differentiation stages and live/dead class labels for organoids under TNF-$\alpha$-induced stress.

We annotated two datasets using NOA's annotation tools. The differentiation stage dataset contains 394 organoids from 2 wells categorized in 4 classes (Spheroid I - 104, Spheroid II - 134, Budding - 119, Enteroid - 37). The viability dataset contains 330 organoids (142 live, 188 dead) from 4 wells with varying TNF-$\alpha$ concentrations.

On both datasets, we trained a collection of classical ML models: SVC, MLP, Random Forest, KNN, AdaBoost, Gaussian Process. As input, we used  automatically calculated features (area, roundness, perimeter, mean and total intensity of grayscale signal for viability; all features including \texttt{skimage.measure.regionprops} for differentiation). Training and evaluation was conducted inside NOA using 10-fold cross-validation. Table~\ref{tab:ml_results} shows that MLP achieves the highest accuracy of 68.56\% for differentiation stage prediction, while Random Forest achieves best result for viability prediction with an accuracy of 72.12\%. Overall, predictive performance appears to depend on the chosen model.

\section{Conclusion and discussion}
\label{sec:conclusion}

NOA significantly advances automated organoid analysis by integrating detection, segmentation, tracking, feature extraction and ML-based prediction into an accessible GUI platform. Validated through real-world biological experiments, NOA demonstrates practical utility for quantifying morphological changes during differentiation and assessing phototoxic effects from multi-channel images.
The modular training pipeline enables researchers to develop custom feature-based prediction models, while manual annotation tools support dataset creation or curation. By combining automation with manual refinement capabilities, NOA bridges the gap between fully automated approaches and the need for domain expertise in organoid analysis. The tool is open-source and available at: \texttt{https://\allowbreak github.com/\allowbreak Meleray/\allowbreak napari-\allowbreak organoid-\allowbreak analyzer}.

In follow-up work we would like to further streamline the integration of new models by modularizing and standardizing the interface for detection, segmentation, and tracking.
Further, we plan to expand the set of built-in feature extraction models with more advanced pre-trained microscopy foundation models.

\section{Acknowledgments}
\label{sec:acknowledgments}

MK acknowledges the financial support from DAAD (German Academic Exchange Service) in the form of a master studies scholarship. 
LG acknowledges support from the Munich School for Data Science (MUDS). 
This work was supported through the Helmholtz Imaging project "AIOrganoid". 
KC and MY are supported by the BMBF 16LW0432, DFG/\allowbreak TRR353-471011418,  DFG/CRC1371-395357507 and the DFG-504986538 grants.

\section{Compliance with Ethical Standards}

This study does not involve data from human and/or animal subjects and no ethical approval is required.

\section{Conflicts of Interest}

The authors have no relevant financial or non-financial interests to disclose.

% ------------------------------------------------------------------------- 
\bibliographystyle{IEEEbib}
\bibliography{refs}

\end{document}